\documentclass[letterpaper,journal]{IEEEtran}
\usepackage[utf8]{inputenc}
\usepackage{array}
\usepackage{mdwmath}
\usepackage{mdwtab}
\usepackage{eqparbox}
\usepackage{url}
\usepackage{cite}
\usepackage{amsmath,amssymb,amsfonts}
\usepackage{algorithmic}
\usepackage{textcomp}
\usepackage{lettrine}
\usepackage{caption}
\usepackage{subcaption}
\usepackage[export]{adjustbox}
\usepackage{amsmath}
\newcommand{\subparagraph}{}
\usepackage{leftidx}
\usepackage{siunitx}
\usepackage{multirow}
\usepackage{balance}
\usepackage[export]{adjustbox}
\usepackage{booktabs}
\usepackage{bbold}

\usepackage{hyperref}

\begin{document}

\title{\LARGE \bf Advanced Situational Graphs for Robot Navigation \\ in Structured Indoor Environments 
}

\author{Hriday Bavle$^{1}$, Jose Luis Sanchez-Lopez$^{1}$, Muhammad Shaheer$^{1}$, \\ Javier Civera$^{2}$ and Holger Voos$^{1}$ 
\thanks{*This work was partially funded by the Fonds National de la Recherche of Luxembourg (FNR), under the projects C19/IS/13713801/5G-Sky, by a partnership between the Interdisciplinary Center for Security Reliability and Trust (SnT) of the University of Luxembourg and Stugalux Construction S.A., by the Spanish Government under Grants PGC2018-096367-B-I00 and PID2021-127685NB-I00 and by the Arag{\'o}n Government under Grant DGA T45 17R/FSE.
For the purpose of Open Access, the author has applied a CC BY public
copyright license to any Author Accepted Manuscript version arising from
this submission.}
\thanks{$^{1}$Authors are with the Automation and Robotics Research Group, Interdisciplinary Centre for Security, Reliability and Trust, University of Luxembourg. Holger Voos is also associated with the Faculty of Science, Technology and Medicine, University of Luxembourg, Luxembourg.
\tt{\small{\{hriday.bavle, joseluis.sanchezlopez, muhammad.shaheer, holger.voos\}}@uni.lu}}%
\thanks{$^{2}$Author is with I3A, Universidad de Zaragoza, Spain
{\tt\small jcivera@unizar.es}}%
}

\maketitle

\begin{abstract}
Mobile robots extract information from its environment to understand their current situation to enable intelligent decision making and autonomous task execution.
In our previous work \cite{s_graphs}, we introduced the concept of Situation Graphs (\textit{S-Graphs}) which combines in a single optimizable graph, the robot keyframes and the representation of the environment with geometric, semantic and topological abstractions. Although \textit{S-Graphs} were built and optimized in real-time and demonstrated state-of-the-art results, they are limited to specific structured environments with specific hand-tuned dimensions of rooms and corridors.

In this work, we present an advanced version of the Situational Graphs (\textit{S-Graphs+}), consisting of the five layered optimizable graph that includes (1) metric layer along with the graph of free-space clusters (2) keyframe layer where the robot poses are registered (3) metric-semantic layer consisting of the extracted planar walls (4) novel rooms layer constraining the extracted planar walls (5) novel floors layer encompassing the rooms within a given floor level. \textit{S-Graphs+} demonstrates improved performance over \textit{S-Graphs} efficiently extracting the room information while simultaneously improving the pose estimate of the robot, thus extending the robots situational awareness in the form of a five layered environmental model. 

\noindent \textbf{Video Link:} \url{https://youtu.be/zPbPe9JXgKk}
\end{abstract}

\section{Introduction}

Recent 3D Scene Graph approaches such as \cite{3d_scene_graph}, \cite{dynamic_scene_graph}, \cite{scene_graph_fusion},  \cite{hydra} model the scene as a graph, in order to efficiently represent the environment and its semantic elements in a hierarchical representation with structural constraints between the elements. Scene graphs enable the robots to understand and navigate the environment similarly to humans, using high-level abstractions (such as chairs, tables, walls) and the inter-connections between them (such as a set of walls forming a room or a corridor). Although these works show promising results, they do not tightly couple scene graphs with SLAM methods that simultaneously optimize the robot poses along with the scene graph. Our previous work \textit{S-Graphs} \cite{s_graphs} bridges this gap by combining a geometric LiDAR SLAM with 3D scene graphs, providing state-of-the-art results. But, \textit{S-Graphs} are only limited to structured environments with specific rooms shapes and size, which caused missed/incorrect room detections in presence of complex room structures leading to incomplete representation of the environment and inaccuracies in the 3D map and robot pose estimate. 

\begin{figure}[t]
    \centering
    \includegraphics[width=0.5\textwidth]{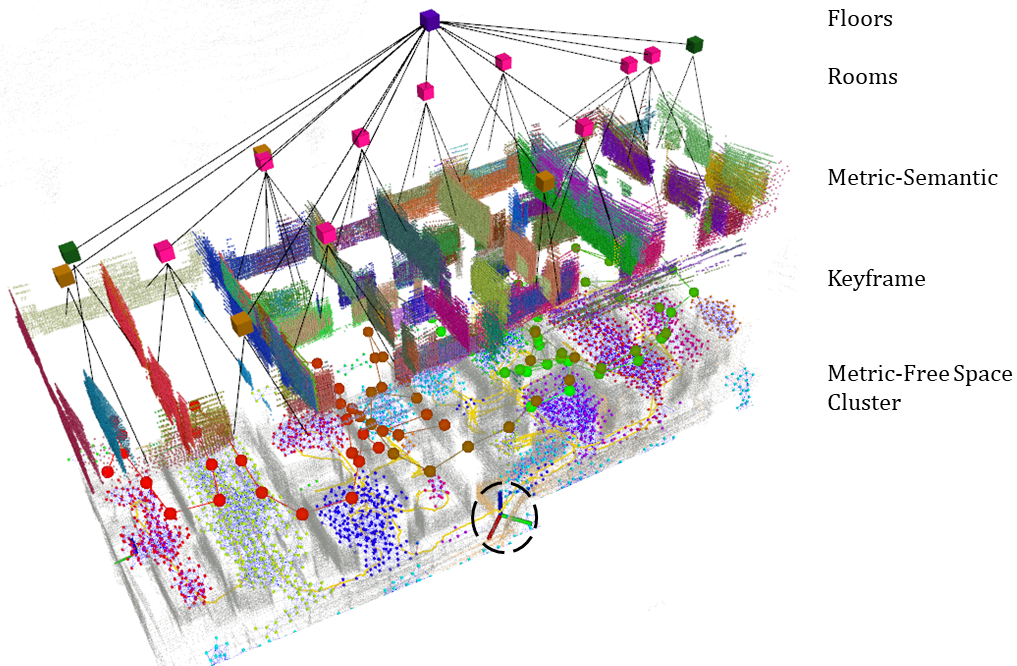}
    \caption{\textit{S-Graph+} generated using a legged robot (circled in black) corresponding to a real construction site consisting of four adjacent houses. The graph can be divided into five sub-layers: 1) Metric-free space cluster layer which consist of the 3D metric-map along with the graph of free space clusters. 2) Keyframe layer registering the robot poses. 3) Metric-Semantic layer extracting planar walls and registering it the corresponding keyframes. 4) Rooms layer constraining the different mapped walls with appropriate room constraints. 5) Floor layer computing the floor centroid using all the mapped walls.}
    \label{fig:scene_graph}
    \vspace{-2mm}
\end{figure}

\begin{figure*}[h]
    \centering
    \includegraphics[width=0.8\textwidth]{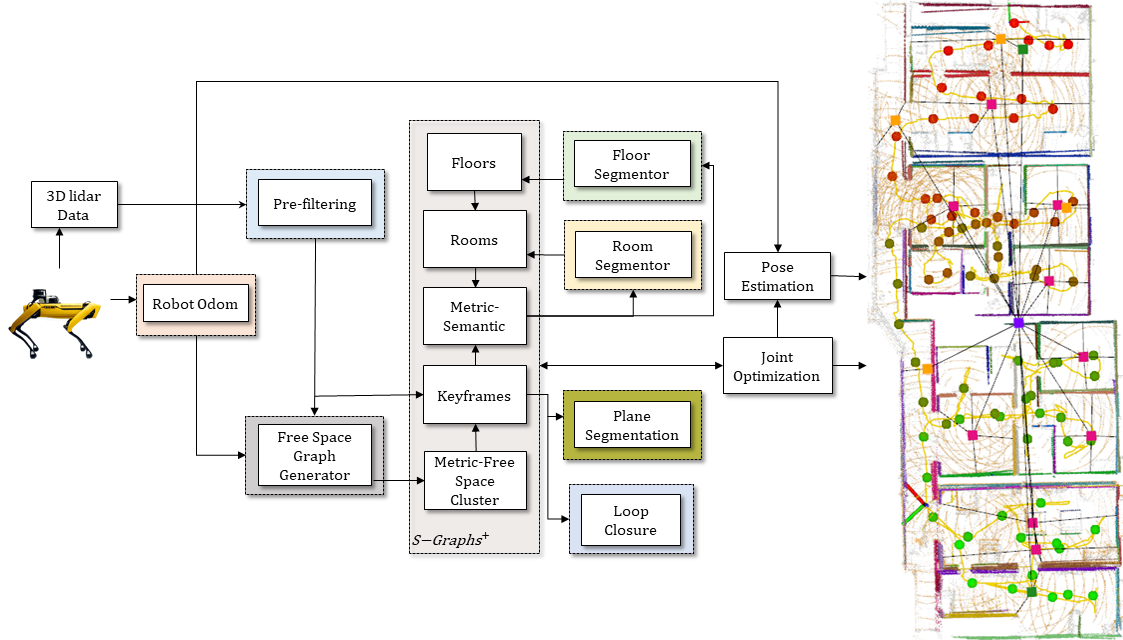}
    \caption{Pipeline of the proposed \textit{S-Graph+} architecture, receiving 3D LiDAR measurements and robot odometry at a given time instant $t$, along with its pre-filtering and free space graph generator. The figure also presents the five layered situational graph along with the loop closure, plane segmentation, room segmentor and floor segmentor modules, which are jointly optimized to update the robot pose and the \textit{S-Graph+}.}
    \label{fig:system_architecture}
    \vspace{-2mm}
\end{figure*}

To overcome the limitations of our previous work, we present \textit{S-Graphs+} (Fig.~\ref{fig:scene_graph}) which contains a five layered optimizable graph comprising of \textit{Metric Layer} consisting of the metric mesh/pointcloud generated using the robot pose and sensor measurements. The metric layer also includes the graph which represents different free space clusters which are generated using a modified version of \cite{topological_graphs}. \textit{Keyframe Layer} consists of the robot poses registered at different time intervals using odometry measurements. \textit{Metric-Semantic Layer} consists of the different planar walls extracted at each keyframe and are constrained to the keyframe using pose-plane constraints. \textit{Rooms Layer} utilizes the graph of free space clusters and the mapped planar walls to identify and map rooms within the given environment. In \textit{S-Graphs+} we define rooms with all four planar walls as finite rooms and rooms with only two opposed planes as infinite rooms. \textit{Floors Layer} is the last layer denoting the floor level in the graph, it utilizes all the mapped planar surfaces to compute the centroid of the current floor level, to then constraint the rooms present at that level. Our main contributions in this paper are: 

\begin{itemize}
    \item A robust and improved implementation of the Situational Graphs named \textit{S-Graphs+} using robot odometry and 3D LiDAR measurements with five hierarchical layers, optimizing the robot poses jointly with a high-level 3D representation of the scene.   
    
    \item The introduction of novel room detection algorithm using the graph of free space clusters and the mapped planar walls, as well as novel factors in the graph to constraint the mapped planar walls with the rooms. 
    \item A thorough experimental evaluation showing the improved performance over the baseline \textit{S-Graph}.  
\end{itemize}

\section{Proposed Approach}
\label{proposed_method}

An overview of the proposed approach is shown in Fig.~\ref{fig:system_architecture}. The overall pipeline can be divided into eight main modules (colored boxes in Fig.~\ref{fig:system_architecture}). The first module pre-filters the LiDAR measurements to remove noise and downsamples the pointcloud. The second module computes the robot odometry either from LiDAR measurements or acquires it from the robot encoders. The third module is a modified version of \cite{topological_graphs} generating free-space graph cluster using the robot poses and 3D LiDAR information.  The fourth module generates the five layered optimizable \textit{S-Graphs+}. \textit{S-Graphs+} utilizes the aid of four additional modules to generate the five layered topological understanding of the environment, namely the plane segmentation module segmenting the planar surfaces from the pointcloud snapshot stored at each keyframe. Room segmentor module utlizing the free-space graph and the mapped planes at a given time instant to detect rooms. Floor segmentor utilizes the information of all the  currently mapped planar surfaces to extract the centroid of the floor node. Finally the loop closure module which utilizes scan-matching algorithm to constraint neighbouring keyframes.

\subsection{Robot Odometry}
As in \textit{S-Graphs} \cite{s_graphs}, we use the Voxelized Generalized Iterative Closest Point (VGICP) \cite{vgicp} for computing the robot odometry using 3D LiDAR measurements, alternatively, as we run our experiments on legged robots, we also test \textit{S-Graphs+} using the odometry estimated from the encoders of these platforms.  

\subsection{Free-Space Graph Generator}
\label{subsection:free_space_graph_generator}
Free-space graph generator is a modified version of \cite{topological_graphs}. In \cite{topological_graphs} a sparse connected graph of free-space is generated given the robot poses and a Euclidean Signed Distance Field (ESDF) map generated using \cite{voxblox}. We divide this connected graph into sub-graphs of places checking the distance of each vertex which respect to obstacles. Vertices which are closer to objects are utilized to disconnect graph into sub-graphs. Using this technique we can obtain disconnected free-space clusters belonging to different rooms, because vertices close to room openings have distances closer to walls (obstacles) and thus vote for disconnecting the graph.   

\subsection{Plane Segmentor}
We use sequential RANSAC to detect all planar surfaces which gives a first estimation of their normals as in \cite{s_graphs}. Compared to \cite{s_graphs}, where plane segmentation is carried out on a different thread which lead to missed planar detections, in \textit{S-Graphs+} the plane segmentation is carried out everytime a new keyframe is registered with the corresponding snapshot of metric 3D LiDAR measurement. This results in efficient detection and mapping of all the planar surfaces at a given time instant.   

\subsection{Room Segmentor} \label{sec:room_segmentor}
Our novel room segmentor utilizes the sub-graphs of free-space clusters (Section.~\ref{subsection:free_space_graph_generator}) and the mapped planar surfaces extracted from the past three keyframes, to detect different rooms with their 2D positions. For each cluster we first check the $l2$ norm between the vertices of each cluster and the points on the mapped planar surfaces. Given two sets of opposed mapped planar walls in $x$ and $y$ axis direction, respectively satisfying a certain width and length and with points closest to the free-space vertices, get identified to form a finite room. We use the four mapped planar walls to calculate the 2D position of our room using the difference between the $x$ axis planes and $y$ axis planes respectively. In cases where only two of the opposed planar walls are identified (either $x$ direction or $y$ direction), the room candidates are called as infinite rooms. To calculate the position of an infinite room with two opposed walls in the direction of the $x$-axis, we utilize the planar coefficients to get its $x$-coordinate while the $y$-coordinate is obtained using the $y$-coordinate of the cluster centroid. The same procedure is applied for infinite rooms with walls facing the $y$-axis. 

\subsection{Floor Segmentor} \label{sec:floor_segmentor}
Floor segmentor utilizes the information from all the currently mapped planar walls with same the floor id to calculate the 2D position of the current floor level the robot is navigating. Whenever the robot ascends or descends to a different floor level, the newly mapped planar walls are incorporated with the new floor id and are the ones used for computing the 2D position of the corresponding floor level.     

\subsection{Loop Closure}
\textit{S-Graphs+} has the same loop closure module as in \cite{s_graphs}, with the room constraints providing the soft loop closure constraints when the robot detects and matches the mapped rooms, while a scan matching-based hard loop closure constraint, constraining neighbouring keyframe poses using their relative pose.  

\subsection{S-Graphs+}

This is the module responsible for creating a five layered hierarchical optimizable graph using the information provided by the above mentioned modules. 

\textbf{Metric-Free-Space Cluster.} This layer contains the 3D metric map of the area generated using the registered keyframe poses of the robot. Bundled within this metric map are the clusters representing the free space within the area.  

\textbf{Keyframes.} This layer creates a factor node $\leftidx{^M}{\mathbf{x}}_{R_t} \in SE(3)$ with the robot keyframe pose at time $t$ in the map frame $M$. The pose nodes are constrained by pairwise odometry readings between consecutive poses. 

\textbf{Metric-Semantic.} This layer creates the factor nodes for the planar surfaces extracted by the planar segmentor. The planar normals extracted in the LiDAR frame $L_t$ at time $t$ are transformed to the global map frame $M$ for its map representation. 
The plane normals with their $\leftidx{^M}{{n}_x}$ or $\leftidx{^M}{{n}_y}$ components greater than the $\leftidx{^M}{{n}_z}$ component are classified as corresponding to vertical planes. Within the vertical planes, those with normals where $\leftidx{^M}{{n}_x}$ is greater than $\leftidx{^M}{{n}_y}$ are classified as $x$-plane normals, and otherwise they are classified as $y$-plane normals. Finally, planes whose normals' bigger component is $\leftidx{^M}{{n}_z}$ are classified as horizontal planes. The planar nodes are constrained with their corresponding keyframes using pose-plane constraints as in \cite{s_graphs}. 

\textbf{Rooms.} The rooms layer receives the extracted room positions with its planar walls from the room segmentor (Section~\ref{sec:room_segmentor}). Finite room node is mapped with edges connecting the four planar walls, and similarly an infinite room node is mapped using edges with two opposed walls. \textit{S-Graphs+} creates novel edges between the room position and each of it planar walls by computing the difference between the vector of plane node coefficients and room position. Data association for the room node is also based on the $l2$ norm. We can safely increase the matching threshold of the rooms, as no two rooms overlap. This allows us to merge planar structures duplicated due to inaccuracies. Similar procedure is carried out for mapping infinite rooms, with either $x$ planes or $y$ planes. 

\textbf{Floors.} The floors layer creates a floor node using the detections from the floor segmentor (Section.~\ref{sec:floor_segmentor}) representing the center of a floor level. Currently, the floor node creates a position-position edges between between all the mapped room at that level. As the estimate of the floor node might change while the robot explores the surrounding, the floor node estimate and its corresponding edges with the rooms are also modified accordingly. This factor results in room nodes remaining bounded within a given floor level.

\section{Experimental Validation}

We validate \textit{S-Graphs+} on datasets generated using both simulated and real-world indoor scenarios, comparing it against several state-of-the-art LiDAR SLAM frameworks and the baseline \textit{S-Graphs}. The datasets are collected teleoperating a Boston Dynamics Spot\footnote{\url{https://www.bostondynamics.com/products/spot}} robot equipped with a Velodyne VLP-16 3D LiDAR. \textit{S-Graph+}  runs real-time on these datasets on-board an Intel i9 16~core workstation.

\subsubsection{Simulated Experiments}
We conduct a total of four simulated experiments. Two of them are performed in environments generated from the 3D mesh of two floors of actual architectural plans provided by a construction company. We denote these two settings as Construction Floor-1 (\textit{CF-1}) and Construction Floor-2 (\textit{CF-2}). We also generated two additional simulated environments resembling typical indoor environments namely Simulated Environment-1 (\textit{SE-1}), and Simulated Environment-2 (\textit{SE-2}). In all the simulated experiments the robot odometry is estimated only from LiDAR measurements. The simulated experiments are performed mainly to validate accuracy of the algorithms with ground truth data using Absolute Trajectory Error (ATE)\cite{evo_traj_calc} due to the absence of ground truth trajectory in real experiments. As can be observed from Table.~\ref{tab:ate_simulated_data}, \textit{S-Graphs+} outperforms its baseline \textit{S-Graphs} \cite{s_graphs} being able to identify and map rooms without requiring fine tuning of the parameters for room identification. Although in experiment \textit{SE-2} the \textit{S-Graphs+} provides second best results, it identifies rooms without any additional parameter adjustment, as was required by \textit{S-Graphs}.

\begin{table}[]
\centering
\caption{Absolute Trajectory Error (ATE) [m], of our \textit{S-Graph+} and several baselines on simulated data. Best results are boldfaced, second best are underlined.}
\begin{tabular}{l | c c c c}
\toprule
& \multicolumn{4}{l}{\textbf{Dataset}} \\
\toprule
\textbf{Method} & \textit{CF-1} &  \textit{CF-2}  & \textit{SE-1} & \textit{SE-2}\\ \midrule
HDL-SLAM \cite{hdl_graph_slam} & 0.09 & {0.11} & {0.04} & {0.15} \\ 
ALOAM \cite{loam} & 0.07 & 0.10 & 0.16 & 0.32 \\
MLOAM \cite{mloam} & 0.15 & 0.39 & 0.65 & 2.82 \\ 
FLOAM\cite{floam} & 3.90 & 0.44 & 0.15 & 0.24 \\ 
SCA-LOAM\cite{sca_loam} & 0.45 & 0.43 & 0.43 & 0.64 \\ 
LeGO-LOAM \cite{lego-loam}  & - & - & - & - \\
\textit{S-Graph - w/o top layer} & {0.05} & {0.17} & {0.40} & {1.01} \\ 
\textit{S-Graph (Baseline)} & \underline{0.04} & \underline{0.07} & \underline{0.03} & \textbf{0.05}\\ 
\textit{S-Graph+} & \textbf{0.03} & \textbf{0.04} & \textbf{0.02} & \underline{0.13}\\ 
\bottomrule
\end{tabular}
\vspace{-3mm}
\label{tab:ate_simulated_data}
\end{table}

\subsubsection{Real Experiments}
We have tested \textit{S-Graphs+} on different real indoor environments but due to space limitations, we provide results with only three real experiments on structured indoor environments ranging from construction site to office environments. We utilize the same robot encoder odometry for all the methods to have a fairer comparison. The first two experiments are performed on two floors of an on-going construction site, the same scenes whose meshes were utilized to validate the algorithm in the simulated environments (\textit{CF-1} and \textit{CF-2}). We also perform a similar experiment in an office environment with a long corridor (\textit{LC-1}) that the robot traverses back and forth. To validate the accuracy of each method on these first three experiments, we report the RMSE of the estimated 3D maps against the actual 3D map generated from the architectural plan. Table~\ref{tab:rmse_real_data} presents the errors obtained in the real experiments and as can be seen, \textit{S-Graphs+} provides better robot pose and map estimation given the accurate identification of rooms and without requiring any parameter tuning. 

\begin{table}[ht]
\caption{Point cloud RMSE [m] on the real datasets. Best results are boldfaced, second best are underlined.}
\centering
\begin{tabular}{l | c c c}
\toprule
& \multicolumn{3}{l}{\textbf{Dataset}} \\
\toprule
\textbf{Method} & \textit{CF-1} &  \textit{CF-2} & \textit{LC-1} \\ \midrule
HDL-SLAM & 1.34 & 0.37 & {1.45}  \\ 
ALOAM  & 8.03 & 1.20 & 3.14  \\ 
MLOAM  & 3.73 & 1.93 & 1.68  \\ 
FLOAM & 7.63 & 1.15 & 2.90  \\
SCA-LOAM & 4.86 & 0.75 & 2.89 \\ 
LeGO-LOAM & 4.08 & 0.70 & 3.40 \\
\textit{S-Graph - w/o top. layer} & {1.21} & \underline{0.36} & {1.45}  \\ 
\textit{S-Graph (Baseline)} & \underline{0.33} & {0.37} & \underline{1.32}  \\ 
\textit{S-Graph+} & \textbf{0.31} & \textbf{0.33} & \textbf{1.31}  \\ 
\bottomrule
\end{tabular}
\vspace{-3mm}
\label{tab:rmse_real_data}
\end{table}

\section{Conclusion}

In this work we present the preliminary version of \textit{S-Graphs+} consisting of a five layered optimizable graph which includes (1) metric layer along with the graph of free-space clusters (2) keyframe layer where the robot poses are registered (3) metric-semantic layer consisting of the extracted planar walls (4) novel rooms layer constraining the extracted planar walls (5) novel floors layer encompassing the rooms within a given floor level. We validated \textit{S-Graphs+} on different simulated and real experiments and observed better performance over its baseline \textit{S-Graphs} with a robust identification of rooms as well as not requiring constant fine tuning of the parameters over all the experimental validation. Our plan is to validate \textit{S-Graphs+} over different real construction scenarios to compare its accuracy with the baseline and other 3D LiDAR slam approaches. 

\balance
\bibliographystyle{IEEEtran}
\bibliography{Bibliography}

\vfill

\end{document}